\documentclass[11pt]{article}

\usepackage[preprint]{acl}

\usepackage{times}
\usepackage{listings}
\usepackage{tcolorbox}
\tcbuselibrary{listings,breakable}

\usepackage[T1]{fontenc}

\usepackage[utf8]{inputenc}

\usepackage{microtype}
\usepackage{hyperref}

\usepackage{inconsolata}
\usepackage{todonotes}
\usepackage{graphicx}
\usepackage{subcaption}
\usepackage{placeins}

\usepackage[table]{xcolor}

\usepackage{booktabs}
\usepackage{multirow}
\usepackage{siunitx}
\usepackage{xcolor}

\title{Towards Reliable Multilingual LLMs-as-a-Judge: An Empirical Study}

\author{
  \textbf{Irune Zubiaga} \quad
  \textbf{Aitor Soroa} \quad
  \textbf{Rodrigo Agerri} \\
  HiTZ Center - Ixa, University of the Basque Country EHU \\
  \texttt{\{irune.zubiaga, a.soroa, rodrigo.agerri\}@ehu.eus}
}

\begin{document}    
\maketitle
\begin{abstract}
Large language models (LLMs) are increasingly used for the automatic evaluation of generated text, yet most prior work focuses on English. Despite the growing demand for multilingual evaluation, extending LLM-based evaluators to multilingual settings remains challenging, particularly for low-resource languages and scenarios where in-domain data is scarce. This work explores several strategies for developing multilingual LLMs-as-a-judge, considering whether in-domain data is available for fine-tuning or not. We systematically analyze English, Spanish, and Basque, representing high-, mid-, and low-resource languages, considering instruction translation, monolingual versus multilingual supervision, and model size. For evaluation, we extend two existing meta-evaluation datasets to Basque and Spanish. Our results reveal key trade-offs: When in-domain data is available, fine-tuned smaller models can achieve performance comparable to proprietary models, whereas zero-shot evaluation with larger models proves more effective in out-of-domain settings. We also observe that fine-tuning on out-of-domain data can adversely affect model performance. These findings provide practical guidance for building efficient, reliable multilingual evaluation pipelines. The data and code are publicly available at \href{https://github.com/hitz-zentroa/mJudge}{hitz-zentroa/mJudge}.
\end{abstract}

\section{Introduction}

Large language models (LLMs) are increasingly used not only for text generation but also for evaluating the outputs of other LLMs. However, most existing research remains centered on English, leaving open research questions unanswered about how to extend LLM-based evaluation to multilingual contexts, especially when low-resource languages are involved.~\footnote{We use low-resource to denote limited representation and performance in current open-weight language models, rather than overall linguistic vitality~\citep{nigatu-etal-2024-zenos}.}

In this work, we study how to design effective training and evaluation strategies that enable robust multilingual performance across languages with different levels of support in existing pretraining corpora. We focus on three languages with varying resource availability: English (high-resource), Spanish (mid-resource), and Basque (low-resource). 
Typologically, they span distinct families and structures: English is Germanic, Spanish is Romance, and Basque is a language isolate with agglutinative morphology. These languages provide a representative and controlled testbed for analyzing how evaluation strategies behave across multilingual conditions, while capturing complementary typological and resource-level variation in a computationally tractable setting. 

We frame our investigation around two conditions: (i) the availability of in-domain data for fine-tuning the evaluators, and (ii) the absence of in-domain training data. Each scenario raises a distinct set of evaluation questions. When training data is available, we examine how translating the data from English to the target language affects performance, the relative benefits of cross-lingual transfer versus multilingual training, and the impact of model size. We also investigate whether it is more effective to train a model directly on target-language data or to leverage a model already proficient in said language and adapt it to the task. Conversely, in the absence of in-domain data, we explore whether fine-tuning on related out-of-domain data is helpful and how model size influences zero-shot generalization.

This work makes several key contributions. 
First, in settings with in-domain data, we find that smaller 8B models remain highly competitive with much larger 70B and proprietary models, suggesting that scale alone is not the decisive factor when task-specific data is available. 
Second, we observe that multilingual training improves performance in Spanish and Basque compared to monolingual training, indicating effective cross-lingual transfer for evaluation tasks. We further find that providing evaluation rubrics and instructions in English, rather than translating them into the target language, yields more consistent judgments across languages, suggesting that full prompt translation is often unnecessary. 
Third, we find that starting from a language-proficient model and adapting it to the evaluation task is generally more effective than training a model on target-language task-specific data.
Finally, in out-of-domain settings, we find that fine-tuning on related data can degrade performance in larger models, while smaller models benefit but still underperform compared to zero-shot large models.


\section{Related Work}

As the use of LLMs continues to grow, the need for multilingual capabilities becomes increasingly important to ensure accessibility for a diverse range of users. While most LLMs remain English-centric, new models targeting other languages have started to emerge. For instance, Pangea~\citep{yue2025pangea} is a fully open multilingual and multimodal LLM trained on diverse instructions in 39 languages, designed to improve cross‑cultural representation and evaluation coverage. Similarly, EuroLLM~\citep{martins2024eurollmmultilinguallanguagemodels} develops a suite of open‑weight multilingual models supporting all official EU languages, demonstrating strong performance on generation and translation benchmarks. Efforts have also focused on broader multilingual coverage, including low-resource languages, such as BLOOM: A 176B-Parameter Open-Access Multilingual Language Model~\citep{workshop2023bloom176bparameteropenaccessmultilingual}. These initiatives highlight the importance of developing and evaluating models that cater to linguistically diverse populations.


Multilingual pre-training can improve cross-lingual transfer to low-resource languages, although gains tend to saturate as model capacity is shared across many languages, a phenomenon known as the curse of multilinguality~\citep{conneau-etal-2020-unsupervised, chang-etal-2024-multilinguality}. Cross-lingual transfer helps mitigate data scarcity~\citep{zoph-etal-2016-transfer, conneau2019cross}, but fine-tuning on additional languages may introduce interference or catastrophic forgetting~\citep{garcia-etal-2021-towards}. To address this, prior work explores selecting related source languages and parameter-efficient methods such as adapter-based approaches, which can improve transfer in low-resource settings~\citep{bapna-firat-2019-simple, pfeiffer-etal-2021-adapterfusion}.

Due to the poor correlation of traditionally used metrics such as BLEU, ROUGE-L or BertScore with human judgments~\citep{sai-survey-2022}, LLMs are increasingly being used as evaluators, achieving considerably higher correlations~\citep{NEURIPS2023_91f18a12}. Unfortunately, most of this work relies on proprietary LLMs~\citep{minaee2024large}, which limits reproducibility and accessibility. Some open alternatives have emerged, such as JudgeLM~\citep{zhu2023judgelm} or Prometheus~\citep{kim-etal-2024-prometheus}. However, most of this research has been conducted primarily in English. While a few works such as M-Prometheus~\citep{pombal2025mprometheussuiteopenmultilingual} have begun exploring multilingual settings, determining the best method for training a multilingual LLM evaluator remains an open and largely unexplored research question.

While some efforts have been made in evaluating these systems, benchmarks remain scarce. A large portion of existing benchmarks are automatically generated using GPT or similar LLMs. While such benchmarks are relatively easy to scale, they can introduce model bias, circularity, and lack of robustness, since the same families of models are often used for both generation and evaluation~\citep{laskar-etal-2024-systematic}. On the other hand, human-annotated benchmarks are more reliable but also more costly and difficult to construct. A considerable amount of these rely on preference-based annotations (e.g., pairwise comparisons between outputs), rather than absolute scoring~\citep{gu2025surveyllmasajudge}. However, prior work shows that pairwise evaluation can amplify biases in LLM-as-a-judge, such as favoring verbosity or authoritative tone, whereas pointwise evaluation, which assesses outputs independently, is less susceptible to these issues~\citep{jeong-etal-2025-comparative}.

Nonetheless, some human-scored resources for absolute evaluation are available for English, such as \texttt{FLASK}~\citep{ye2024FLASK}, highlighting the potential value of expanding this line of work. However, very few available multilingual benchmarks are currently available. Among them, \texttt{RECON}~\citep{Doddapaneni_Khan_Venkatesh_Dabre_Kunchukuttan_Khapra_2024} stands out as a human-supervised, multilingual evaluation dataset. It contains prompts written in multiple languages paired with English references and evaluation rubrics. However, the currently covered languages (Bengali, German, French, Hindi, Telugu, and Urdu) remain limited. While this set spans both high- and low-resource languages, the scope of \texttt{RECON} is still narrow, highlighting the need for broader multilingual benchmarks for meta-evaluation. Expanding such resources is crucial for enabling more comprehensive assessment and comparison of LLMs across diverse languages.

\section{Dataset Creation}


Since no publicly available judgment datasets exist for Basque or Spanish, we rely on existing English datasets and adapt them to these languages. Each instance in both train and test data consists of an instruction, a candidate output, an evaluation rubric, a reference answer, and a scalar score ranging from 1 to 5, along with an explanation of the assigned score. Each component is described in more detail in Appendix~\ref{feedbackdataset}.

\subsection{Training corpus}~\label{training_data}
We automatically translate the \texttt{FeedBack Collection}\footnote{\url{https://huggingface.co/datasets/prometheus-eval/Feedback-Collection}} into Basque and Spanish obtaining a parallel dataset\footnote{We translate text into Spanish using \href{https://huggingface.co/meta-llama/Llama-3.1-70B-Instruct}{Llama-3.1-70B-Instruct} and into Basque using \href{https://huggingface.co/HiTZ/Latxa-Llama-3.1-70B-Instruct}{Latxa-Llama-3.1-70B-Instruct}.}. The corpus, originally introduced in \citet{preometh}, consists of 100k instruction–response pairs annotated with explicit evaluation instructions, evaluator feedback, and scalar scores. The translation prompt is given in Appendix~\ref{sec:prompt}. The dataset was selected due to its wide validation in prior work \citep{kim-etal-2024-prometheus, pombal2025mprometheussuiteopenmultilingual} and its explicit rubric-based scoring, which helps ground model evaluations and produce more reliable assessments~\citep{hashemi-etal-2024-llm}. Following common practice in LLM-based translation, we use a temperature of $T=0.2$, while all other settings were kept at their default values. This low temperature reduces stochasticity, producing more consistent and reproducible outputs. We retain the original training and validation splits.

\subsection{Evaluation benchmarks}\label{sec:eval}
In an effort to extend multilingual evaluation coverage, we translate two general instruction-following English benchmarks: \texttt{RECON}~\citep{Doddapaneni_Khan_Venkatesh_Dabre_Kunchukuttan_Khapra_2024} and \texttt{FLASK}~\citep{ye2024FLASK}. We treat \texttt{RECON} as in-domain, as it follows the same generation protocol as the training data described in \citet{preometh}. In contrast, \texttt{FLASK} is considered out-of-domain, as it is based on a different annotation paradigm relying on human scoring of model outputs.

\texttt{RECON} is a meta-evaluation benchmark consisting of 500 instances. Similarly to the \texttt{FeedBack Collection} dataset, it was created by first collecting 500 original instructions and then prompting GPT-4o to generate question-specific evaluation criteria and detailed scoring rubrics. Evaluation responses are subsequently generated by prompting GPT-4o to produce answers targeting each score $x \in \{1,\ldots,5\}$ according to these rubrics. Although the benchmark is already multilingual, it does not originally include Spanish or Basque; by translating it into these languages, we broaden its linguistic coverage and create additional partitions that may be valuable for future multilingual evaluation research. 

In contrast, \texttt{FLASK} (Fine-grained Language Model Evaluation based on Alignment Skill Sets) is a 2000-instance benchmark that serves as an out-of-domain evaluation set, annotated entirely by humans. It provides fine-grained judgments across multiple alignment dimensions, including helpfulness, safety, and factuality. For each instance, three independent human scores are provided. We use the mode of these scores as the final label.

The translation setup follows that used for the training corpus (Section~\ref{training_data}), employing the same models and hyperparameters for machine translation. 
\texttt{RECON} was further post-edited by professional translators, who were instructed to ensure both correctness and naturalness, resulting in a human-verified reference subset. We then use this data to assess whether machine translation alone is sufficient, or whether post-editing is necessary for accurate performance estimation, by evaluating the statistical significance of differences between the MT and POS benchmarks (see Appendix~\ref{posvsmt}). We found that the models did not differ systematically in their sensitivity to translation quality. These findings indicate that the benchmarks do not substantially affect the relative performance of the evaluated models, suggesting that the MT dataset is suitable for identifying the most effective training strategies for multilingual LLMs-as-a-judge. Consequently, we retain the machine-translated version of \texttt{FLASK}. We provide further analysis of translation quality in Appendix~\ref{mt_quality}.

\section{Experimental Setup}

In our experiments, we consider two evaluation scenarios. In the first scenario (see Section~\ref{sec:taskdata}), in-domain task-specific data is available to train the LLM-based judges. In the second scenario (see Section~\ref{sec:notaskdata}), no in-domain data is available, reflecting a setting commonly encountered in practice where only related out-of-domain data may be accessible. Our experimental setting is designed to analyze how multilingual tasks can be evaluated under each of these conditions and to identify the most effective evaluation strategies in each case.

We assess models in-domain with \texttt{RECON} and out-of-domain with \texttt{FLASK}, using the evaluation prompts and scoring rubrics provided by each benchmark. In both cases, we train the judge models under different language configurations to study how models and training languages influence multilingual evaluation performance. Using the parallel data described in Section~\ref{training_data}, we consider two training data configurations: monolingual, where models are trained on a single language (Basque, Spanish, or English), and multilingual (dubbed \emph{multi}), where parallel instances across all three languages are combined, resulting in a dataset three times larger. The combined dataset is then shuffled to ensure a balanced and randomized distribution of examples.

\subsection{Model choice}~\label{sec:model_choice}
For our analysis, we selected models from the LLaMA family~\citep{touvron2023llama}, as they support English, Spanish, and Basque to varying degrees and comprise a diverse set of models with varying capabilities and scales. 

Alongside \texttt{Llama\allowbreak-3.1\allowbreak-8B\allowbreak-Instruct}, we evaluate two LLaMA 3.1-based variants: \texttt{DeepSeek\allowbreak-R1\allowbreak-Distill\allowbreak-Llama\allowbreak-8B}~\citep{deepseekai2025deepseekr1incentivizingreasoningcapability}, fine-tuned on reasoning data, and \texttt{Latxa\allowbreak-Llama\allowbreak-3.1\allowbreak-8B\allowbreak-Instruct}~\citep{sainz-etal-2025-instructing}, fine-tuned and instructed for Basque, and therefore proficient in the language. Our main analysis uses the 8B variants, while the 70B versions of Llama-Instruct and Latxa are additionally evaluated to analyze scaling effects and whether larger models yield substantial gains over smaller ones.
The selected models represent three scenarios: an instruction-tuned model, a model fine-tuned on reasoning data, and a model already proficient in the target language. 

We fine-tune the models using the llama-recipes\footnote{\url{https://github.com/facebookresearch/llama-recipes}} framework, updating all model parameters. Models are trained for 3 epochs with a learning rate of $1e^{-5}$, which was selected via grid search, and the context length is set to 8,100 tokens to accommodate the maximum length of inputs in the Basque dataset. We use a batch size of 1 with gradient accumulation over 4 steps. Fully Sharded Data Parallel (FSDP) is enabled with bfloat16 precision, and fast kernels are used to improve training efficiency. During training, validation loss is monitored and the checkpoint with the lowest validation loss is selected for evaluation.

To guarantee scoring consistency, we run all models with deterministic decoding ($T=0$) at inference time, which eliminates stochastic variation in the assigned scores and ensures that the evaluations reflect the models’ rating behavior rather than sampling noise. We use constrained decoding to enforce the required output format (Appendix~\ref{reg_pat}) and set the maximum generation length to match the training context length. Inference is performed using the vLLM framework~\citep{kwon2023efficient}, which manages batching automatically and optimizes GPU utilization.

Given the computationally and memory-intensive nature of this work, all experiments were conducted on high-performance computing (HPC) systems to ensure efficient execution at scale.

\section{Results}
We organize our results across two evaluation scenarios: when in-domain training data is available and when it is not. The subsections are further structured around our key research questions, highlighting how data availability influences evaluation outcomes.

\subsection{When In-Domain Training Data Is Available}\label{sec:taskdata}
We train on the feedback dataset and evaluate on \texttt{RECON}, which serves as an in-domain benchmark for the task. This setting isolates the effect of model design choices under ideal supervision conditions.

\paragraph{Must evaluation instructions be translated into the target language, or do English-only instructions suffice when evaluating multilingual tasks?}

To answer this question, we consider two different data translation scenarios:
\begin{itemize}
    \item \textit{Full translation (\texttt{lang})}: In this configuration, the entire dataset—including input prompts, reference answers, scoring rubrics, and model outputs—is translated into the target language. The resulting corpora are denoted as \texttt{en}, \texttt{es}, \texttt{eu}, and \texttt{multi}, corresponding to English, Spanish, Basque, and multilingual, respectively.
    \item \textit{Partial translation (\texttt{io\_lang})} : In this configuration, only the original instructions and model outputs are translated, while all other components (such as scoring rubrics and system prompts) remain in English. These variants are denoted as \texttt{io\_en}, \texttt{io\_es}, \texttt{io\_eu}, and \texttt{io\_multi}.
\end{itemize}

This comparison assesses whether translating the evaluation rubrics provides measurable benefits or if keeping them in English suffices, considering both training and test to capture potential inference-time effects.

\begin{figure}[htb]
    \centering
    \includegraphics[width=0.9\linewidth]{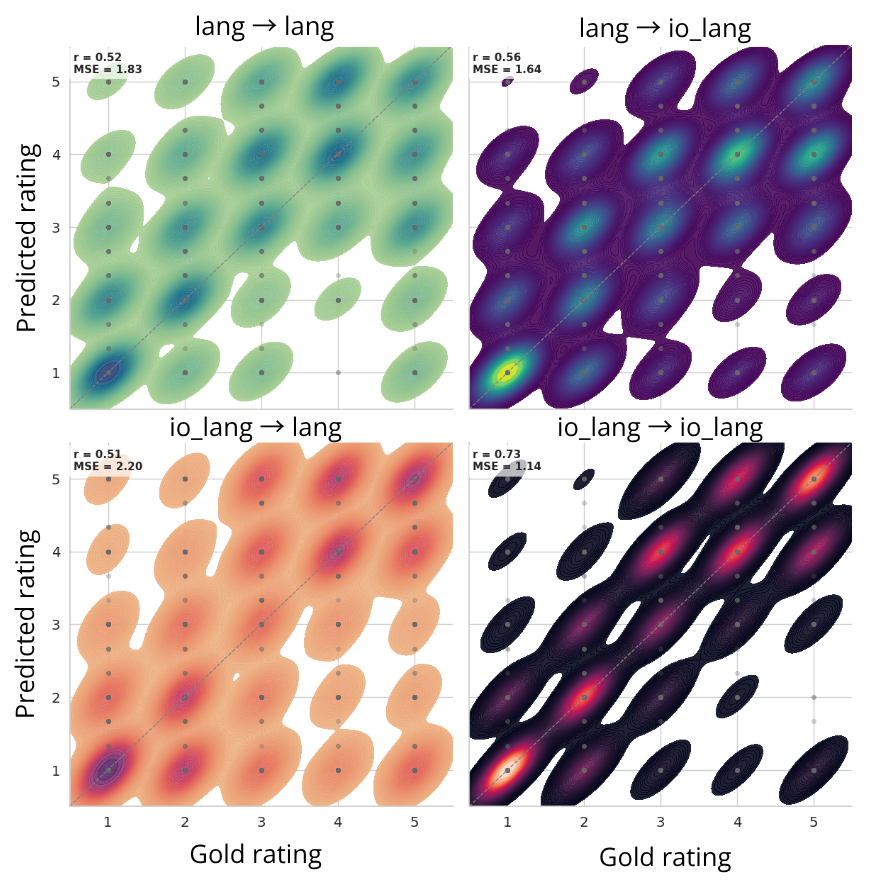}
    \caption{Density plot of predictions against gold labels on the in-domain \texttt{RECON} benchmark. We denote configurations as $X \rightarrow Y$, where $X$ indicates the training-time language setup and $Y$ the evaluation-time language setup. \texttt{lang} refers to fully translated instructions (\texttt{es}, \texttt{eu}, or \texttt{multi}), and \texttt{io\_lang} refers to partially translated instructions, where only the original instructions and model outputs are translated while the rest is kept in English.}
    \label{fig:recondensityplot}
\end{figure}
Figure~\ref{fig:recondensityplot} visualizes model predictions against gold labels. A key observation is the substantial gain obtained when evaluating on partially translated benchmarks: models evaluated with English instructions ($X \rightarrow io\_lang$) consistently outperform those evaluated with fully translated instructions ($X \rightarrow lang$). Our results show that using English-only evaluation instructions at test time is not only sufficient but often preferable for multilingual evaluation, even when the underlying model is trained monolingually.

In this test-time setting, models trained on fully translated data ($X = lang$) exhibit weaker alignment with reference judgments, whereas models trained on partially translated data ($X = io\_lang$) show substantially stronger alignment. In the latter setting, the correlation with gold labels reaches 0.73, compared to 0.56 for monolingual fine-tuning. The density plot makes this improvement clear: predictions concentrate tightly along the diagonal with reduced off-diagonal spread, indicating better calibration and greater consistency with gold labels resulting in a lower mean squared error (MSE). We also observe a slight tendency across settings toward score overestimation rather than underestimation, as the region above the diagonal appears denser than the region below it. Together, these results suggest that retaining English evaluation rubrics during training in a different target language is beneficial for model performance. 

Based on these findings, we adopt English evaluation instructions (\texttt{io\_lang}) as the standard training and test-time configuration in all subsequent experiments. For notational simplicity, \textbf{\texttt{eu} and \texttt{es} will henceforth be used to refer to \texttt{io\_eu} and \texttt{io\_es}, respectively}.


\paragraph{Multilingual training and cross-lingual transfer}


We analyze whether multilingual training provides advantages over monolingual models, and to what extent models trained on one language can be applied to other languages. To this end, we train models separately on each language (English, Spanish, or Basque) and evaluate their performance on the remaining languages to measure cross-lingual transfer. We also compare their performance to multilingual variants trained on the combined parallel dataset. 
We report the mean of each metric across model variants in Table~\ref{tab:mean_language}. Individual scores per model are reported in Appendix~\ref{ap:multiandcross}.

\begin{table}[htb]
\small
\centering
\setlength{\tabcolsep}{5pt}
\resizebox{\columnwidth}{!}{%
\begin{tabular}{|l|c|c|c|c|c|c|}
\cline{2-7}
 \multicolumn{1}{c|}{} & \multicolumn{2}{c|}{\texttt{RECON} ${en}$} & \multicolumn{2}{c|}{\texttt{RECON} ${es}$} & \multicolumn{2}{c|}{\texttt{RECON} ${eu}$} \\
\cline{2-7}
\multicolumn{1}{c|}{}& r $\uparrow$ & MSE $\downarrow$ & r $\uparrow$ & MSE $\downarrow$ & r $\uparrow$ & MSE $\downarrow$\\
\hline
     en &  \cellcolor{gray!30}\textbf{0.840} &\cellcolor{gray!30}\textbf{0.773} &0.663&2.131&0.636&1.693\\
     es &0.767&1.059& \cellcolor{gray!30}0.771 &  \cellcolor{gray!30}1.071 &0.671&1.566 \\
     eu &0.755&1.091&0.757& 1.117 &  \cellcolor{gray!30}0.724 &  \cellcolor{gray!30}1.201\\
     \hline\hline
    multi & 0.793 & 0.919 &  \textbf{0.785} &  \textbf{0.961} & \textbf{0.766} & \textbf{1.029}\\
    \hline\hline
    zs & 0.602 & 1.693 & 0.468& 2.344 & 0.567 & 2.05\\
\hline
\end{tabular}
}
\caption{Mean model performance per language on \texttt{RECON}. The models that have been trained and tested on the same language appear in gray. The best score in each language appears in bold.}
\label{tab:mean_language}
\end{table}

We first focus on comparing multilingual and monolingual settings when the training and test languages are the same.
Multilingual training yields improvements for Basque ($4.2$ points in Pearson) and Spanish ($1.4$ points in Pearson), while performance degrades in English (a drop of $4.7$ points in Pearson). These results corroborate previous findings that cross-lingual transfer mostly occurs from high-resource languages to lower-resource ones. In the cross-lingual setting, results indicate that training and testing on the same language remains preferable, as expected. Nevertheless, cross-lingual training consistently outperforms the zero-shot approach, which achieves lower overall performance. This demonstrates that cross-lingual transfer occurs: learning the evaluation task in one language helps the model generalize to others.


\paragraph{The effect of scale}

We analyze the effect of scale by including results for the 70B versions of Llama-Instruct and Latxa in both the multilingual setting (which achieved the best average results for the 8B models) and the zero-shot setting. For baseline comparison, we include results from available open models (the Prometheus family) and proprietary models (GPT-5.2 and Command-A).


\begin{table}[ht]
\small\centering
\setlength{\tabcolsep}{2pt}
\begin{tabular}{|l|c|c|c|c|}
\hline
Model & \texttt{RECON} ${en}$ & \texttt{RECON} ${es}$ & \texttt{RECON} ${eu}$\\
 & {\tiny$(\mathrm{r}\uparrow/\mathrm{MSE}\downarrow)$}& {\tiny$(\mathrm{r}\uparrow/\mathrm{MSE}\downarrow)$} & {\tiny$(\mathrm{r}\uparrow/\mathrm{MSE}\downarrow)$}\\
\hline
\hline
Latxa-Inst-8B$_{ft}$ & \underline{0.836/0.724} & 0.816/0.824 & 0.805/0.888 \\
Latxa-Inst-8B$_{zs}$ & 0.526/1.978 & 0.266/3.147 & 0.745/1.042 \\
Latxa-Inst-70B$_{ft}$ & 0.831/0.720 & \underline{0. 816/0.800} & \underline{\textbf{0.828/0.754}}\\
Latxa-Inst-70B$_{zs}$ & 0.803/0.871 & 0.705/1.238 & 0.750/1.062\\
\hline
\hline
Llama-Inst-8B$_{ft}$& 0.741/1.102 & 0.757/1.044 & 0.732/1.140 \\
Llama-Inst-8B$_{zs}$ &0.678/1.408& 0.669/1.540& 0.388/3.058\\
Llama-Inst-70B$_{ft}$ & 0.790/0.876 & 0.754/1.060 & 0.774/0.980\\
Llama-Inst-70B$_{zs}$ &0.812/0.788&0.735/1.074&0.734/1.096\\
\hline
\hline
Prometheus-v1 & 0.828/0.718 & 0.752/1.018 & 0.269/4.894 \\
Prometheus-v2 & 0.826/0.742&0.766/0.964 & 0.588/1.718\\
M-Prometheus & 0.829/0.698 & 0.784/0.844 & 0.689/1.414 \\
\hline
\hline
GPT-5.2	& \textbf{0.844/0.650} & \textbf{0.818/0.770} &  0.700/0.900\\
Command-A &0.777/0.914 & 0.703/1.256 & 0.662/1.436\\
\hline
\end{tabular}
\caption{Comparison with state-of-the-art models using our best-performing setting, including zero-shot evaluation. The best overall model appears in bold and the best open model underscored.}\label{tab:idperformance}
\end{table}

Table~\ref{tab:idperformance} shows that, in zero-shot settings, the models exhibit substantial performance gains as model size increases, with the exception of Latxa on Basque, where performance remains largely stable across scales. However, after fine-tuning, the benefits of scaling are markedly reduced: larger models yield only marginal and inconsistent improvements across evaluation metrics, resulting in limited overall gains relative to the increase in parameter count.
Overall, our models outperform the open baselines and are competitive with proprietary models, with strong improvements in performance for Basque.
These results suggest that, while scaling model size yields limited benefits for this task, targeted fine-tuning can meaningfully improve performance, particularly in languages where the zero-shot baseline is weaker.

\paragraph{Training data language vs. model language proficiency}


We now investigate whether it is more beneficial to adapt the training data to the target language or to leverage a model proficient in that language for the task, focusing our analysis on Basque. Specifically, we compare two strategies: (i) fine-tuning a general-purpose model (Llama-Instruct) on Basque instructions translated from the English corpus, and (ii) fine-tuning a Basque-proficient model (Latxa) using the original English instructions. 
\begin{table}[htb]
\small
    \centering
    \footnotesize
    \begin{tabular}{|c|c|c|c|}
    \hline
         \textbf{Model} & \textbf{train} & \multicolumn{2}{c|}{\texttt{RECON} ${eu}$} \\
        \cline{3-4}
        &\textbf{lang}& $r$ $\uparrow$ & MSE $\downarrow$\\
         \hline
         Latxa-Inst-8B & en & \textbf{0.760} & \textbf{1.065}\\
         Llama-Inst-8B & eu & 0.717 & 1.179\\
    \hline
    \end{tabular}
    \caption{Performance of LLMs-as-a-judge in Basque under two language-alignment strategies: English instructions with a Basque-proficient model vs. Basque-translated instructions with an English model. The best scores appear in bold.}
    \label{tab:modelvstrain}
\end{table}

Table~\ref{tab:modelvstrain} shows the results of both strategies when tested on Basque. The optimal strategy is to fine-tune the Basque-proficient model using English instructions, with a substantial gain of $4.3$ points in Pearson.

\subsection{When In-Domain Training Data Is Unavailable}\label{sec:notaskdata}


We now focus on the out-of-domain setting.
We replicate the analysis from Section~\ref{sec:taskdata}, and evaluate models fine-tuned on the feedback dataset on \texttt{FLASK}, an out-of-domain benchmark.
As in the previous setting, we begin by analyzing the impact of full versus partial translation in the train and test data. 

\begin{figure}[htb]  
    \centering
    \includegraphics[width=0.9\linewidth]{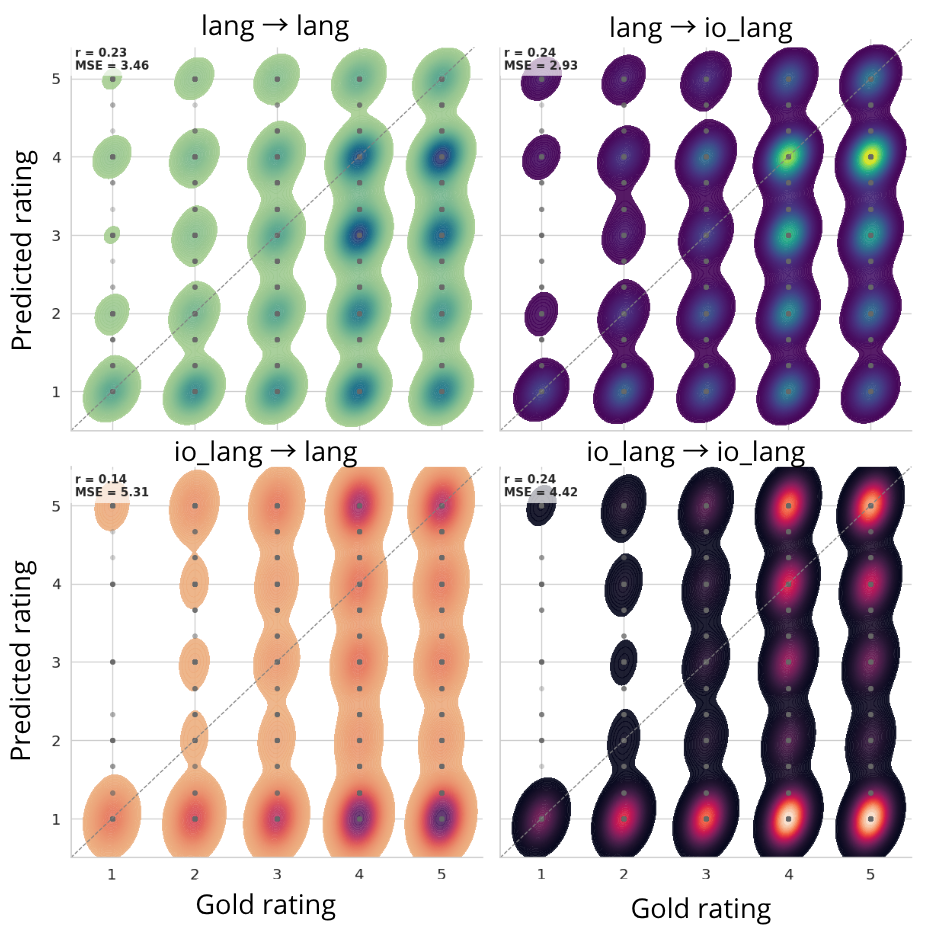}
        \caption{Density plot of predictions against gold labels on the out-of-domain \texttt{FLASK} benchmark. We denote configurations as $X \rightarrow Y$, where $X$ indicates the training-time language setup and $Y$ the evaluation-time language setup. \texttt{lang} refers to fully translated instructions (\texttt{es}, \texttt{eu}, or \texttt{multi}), and \texttt{io\_lang} refers to partially translated instructions, where only the original instructions and model outputs are translated while the rest is kept in English.}
    \label{fig:flaskdensityplot}
\end{figure}

Figure~\ref{fig:flaskdensityplot} shows that the impact of language configuration under out-of-domain evaluation differs substantially from what is observed in-domain. While training on partially translated data and testing on fully translated data (\texttt{$io\_lang \rightarrow lang$}) performs worst, correlation differences across the remaining configurations are marginal, and no single training strategy clearly dominates. Instead, the main differences emerge in the shape of the score distributions and are reflected in the MSE.

Models fine-tuned with English evaluation instructions (\texttt{$X = io\_lang$}) tend to produce more polarized judgments, assigning a higher proportion of extreme scores (1 and 5) and fewer intermediate values. This effect becomes increasingly pronounced at higher score levels: while predictions for score 1 remain largely unaffected, deviations begin to appear at score 3 and become substantial for scores 4 and 5, where extreme predictions dominate. Although this polarization persists when models are trained with fully translated evaluation instructions (\texttt{$X = lang$}), it is notably less pronounced, as reflected by a lower MSE (2.96 compared to 4.42 when testing on \texttt{io\_lang}, and 3.46 compared to 5.31 when testing on \texttt{lang}).

Despite these differences, overall performance remains poor in terms of both Pearson and MSE, indicating limited generalization beyond the training domain. Taken together, these results suggest that the models lack sufficient robustness for out-of-domain multilingual evaluation, and that gains observed in-domain do not reliably transfer across datasets or domains. 

\paragraph{The effect of scale}
\begin{table*}[htb]
\small
\centering
\begin{tabular}{|ll|cc|cc|cc|}
\cline{3-8}
 \multicolumn{1}{c}{}& \multicolumn{1}{c|}{} & \multicolumn{2}{c|}{\textbf{en} {\tiny$(\mathrm{r}\uparrow/\mathrm{MSE}\downarrow)$}} & \multicolumn{2}{c|}{\textbf{es} {\tiny$(\mathrm{r}\uparrow/\mathrm{MSE}\downarrow)$}} & \multicolumn{2}{c|}{\textbf{eu} {\tiny$(\mathrm{r}\uparrow/\mathrm{MSE}\downarrow)$}} \\
\cline{3-8}
\noalign{\vskip 5pt}
\cline{3-8}
  \multicolumn{1}{c}{}&  \multicolumn{1}{c|}{}& zs & ft & zs & ft & zs & ft \\
\hline
\textbf{8B} 
                &\texttt{RECON}& 0.526/1.978 & \cellcolor{gray!30}\textbf{0.841/0.716}& 0.501/2.272 & \cellcolor{gray!30}\textbf{0.723/1.164} & 0.460/2.541 & \cellcolor{gray!30}\textbf{0.735/1.070}\\
                &\texttt{FLASK}& 0.301/3.161 & \cellcolor{gray!30}\textbf{0.393/3.840}& 0.193/2.935 & \cellcolor{gray!30}\textbf{0.258/4.183}& 0.213/3.352 & \cellcolor{gray!30}\textbf{0.308/3.042}\\
\hline
\textbf{70B} &\texttt{RECON}& 0.803/0.871 & \cellcolor{gray!30}\textbf{0.900/0.470}& 0.705/1.238 & \cellcolor{gray!30}\textbf{0.856/0.657}& 0.750/1.062 & \cellcolor{gray!30}\textbf{0.831/0.713}\\
                &\texttt{FLASK}& \cellcolor{gray!30}\textbf{0.594/1.476} & 0.440/3.016& \cellcolor{gray!30}\textbf{0.476/1.657} & 0.350/2.680 & \cellcolor{gray!30}\textbf{0.424/2.502} & 0.336/2.077\\
\hline
\end{tabular}
\caption{Performance of Latxa 70B and 8B on in-domain and out-of-domain settings after finetuning on the multilingual dataset. Best results in bold and gray.}
\label{tab:size_effect}
\end{table*}
Given the limited out-of-domain generalization observed so far, a natural question is whether increasing model size can mitigate these effects. Larger language models are often assumed to generalize better, and we therefore investigate whether fine-tuning larger models leads to more robust evaluation behavior. To this end, we evaluate the previously fine-tuned 70B-parameter Latxa model and compare its performance to its 8B-parameter counterpart. 
Results across languages are summarized in Table~\ref{tab:size_effect}, which also includes in-domain results for comparison.

The table reveals a clear divergence between small and large models: fine-tuning improves both in-domain and out-of-domain performance for smaller models, whereas for 70B-parameter models it enhances in-domain performance but degrades out-of-domain generalization. 
\begin{figure}[htb]
    \centering
    \includegraphics[width=0.9\linewidth]{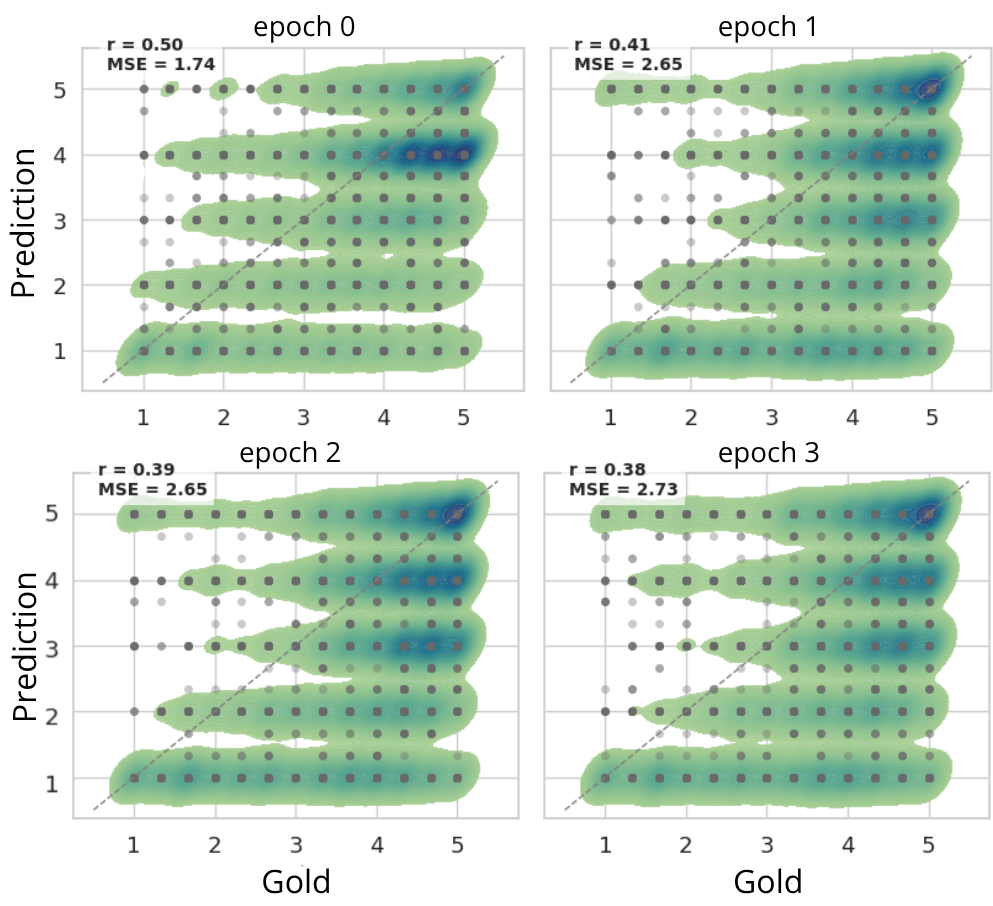}
    \caption{Density plot showing the distribution of predictions produced by the Latxa 70B model after each fine-tuning epoch across the English (en), Spanish (es), and Basque (eu) benchmarks. Epoch 0 corresponds to the zero-shot performance of the model.}
    \label{fig:onsize}
\end{figure}

To further investigate the source of the performance drop after fine-tuning the 70B model, we examine the model’s score distributions across training epochs. As illustrated in Figure~\ref{fig:onsize}, the model tends to assign lower scores than the gold ratings in general. However, after each fine-tuning epoch, we observe a growing concentration of predictions at the maximum score (5).

Taking all of this into account, a plausible explanation for the observed behavior in Table~\ref{tab:size_effect} is that smaller models lack sufficient prior knowledge to perform the task reliably and therefore benefit more from additional supervision. In contrast, larger models already exhibit strong baseline capabilities, and fine-tuning seems to introduce a tendency to systematically assign higher scores than the ground truth, which becomes increasingly pronounced with additional training epochs. This behavior suggests a potential reduction in generalization ability, possibly due to overconfidence induced by fine-tuning.


As a final step, we evaluate several large language models on the \texttt{FLASK} benchmark (see Table~\ref{tab:idperformanceflask}).
\begin{table}[htb]
\small\centering
\setlength{\tabcolsep}{2pt}
\begin{tabular}{|l|c|c|c|c|}
\hline
Model & \texttt{FLASK} ${en}$ & \texttt{FLASK} ${es}$ & \texttt{FLASK} ${eu}$\\
 & {\tiny$(\mathrm{r}\uparrow/\mathrm{MSE}\downarrow)$}& {\tiny$(\mathrm{r}\uparrow/\mathrm{MSE}\downarrow)$} & {\tiny$(\mathrm{r}\uparrow/\mathrm{MSE}\downarrow)$}\\
\hline
\hline
Llama-Inst-70B$_{zs}$ & \underline{0.531/2.035} & \underline{0.450/2.178} & \underline{0.417/2.073} \\
Latxa-Inst-70B$_{zs}$ & 0.524/1.969 & 0.4213/2.236 & 0.372/2.625 \\
\hline
\hline
prometheus-v2 & 0.477/1.944 & 0.413/2.110 & 0.318/3.077\\
prometheus-v1 & 0.275/3.520 & 0.266/2.971 & 0.079/6.673 \\
M-prometheus&0.339/3.520&0.356/3.376&0.227/4.404\\
\hline
\hline
GPT-5.2 &\textbf{0.684/1.372}&\textbf{0.577/1.423} & \textbf{0.606/1.531} \\
Command-A &0.570/1.894 & 0.424/2.399 & 0.436/2.074 \\
\hline
\end{tabular}
\caption{Comparison of the zs performance of the LLaMA models and state-of-the-art models. The best model appears in bold and the best open model underscored.}\label{tab:idperformanceflask}
\end{table}
For each benchmark, we report our best fine-tuned model since no best strategy was identified. Overall, the table shows that zero-shot performance is competitive, with performance generally improving as model size increases.

Taken together, these results suggest that in out-of-domain settings, using large language models in a zero-shot configuration is preferable to task-specific fine-tuning on out-of-domain data.

\paragraph{The effect of finetuning}

\begin{figure*}[htb]
    \centering
    \includegraphics[width=1\linewidth]{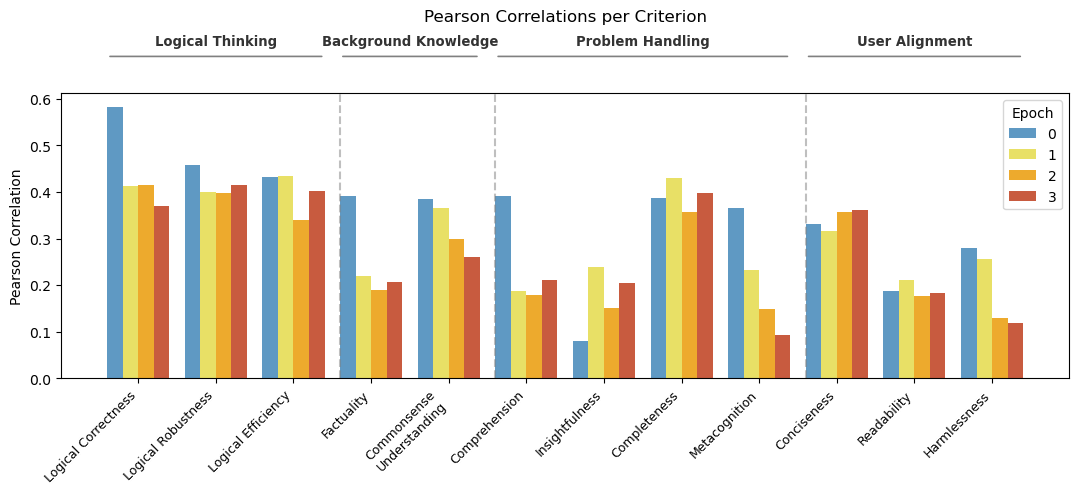}
            \caption{Pearson correlations between model predictions and human judgments across \texttt{FLASK} evaluation criteria after 0, 1, 2 and 3 training epochs.}
    \label{fig:pearson_per_criterion}
\end{figure*}
To further examine the effects of fine-tuning and leverage the categorical structure of \texttt{FLASK}, Figure~\ref{fig:pearson_per_criterion} reports the Pearson correlations for each evaluation criterion using the Latxa 70B model in zero-shot, as well as after fine-tuning for one, two, and three epochs. As shown in the figure, fine-tuning yields modest improvements in some User Alignment aspects, such as Conciseness and Readability, and also slightly improves Insightfulness. However, performance generally declines across other dimensions, particularly those related to Logical Thinking and Background Knowledge, with notable drops in Logical Correctness and Factuality. We also observe a drop in Metacognition, a measure of how well the model avoids responding to ambiguous or uncertain instructions and explicitly acknowledges the uncertainty of its responses, disclosing its limitations when it lacks sufficient information for a reliable answer. The observed decrease in Factuality and Metacognition could be attributed to effects reported in prior work suggesting that fine-tuning can exacerbate hallucinations~\cite{gekhman-etal-2024-fine}. We also observe a noticeable decline in Harmlessness consistent with prior work showing that fine-tuning can reduce safety~\cite{ICLR2024_83b7da3e}. Overall, these results suggest that while fine-tuning may enhance certain stylistic properties, zero-shot performance remains more reliable for knowledge- and reasoning-intensive dimensions.



\section{Conclusions and Future Work}
In this work, we study several strategies for building reliable multilingual LLMs-as-a-judge when natural training data in the target language is unavailable, considering both scenarios where in-domain data is available and where it is not. Focusing on English, Spanish, and Basque, our experiments reveal several practical insights. When fine-tuning data is available, providing evaluation rubrics and instructions in English leads to more consistent judgments across languages, suggesting that translating evaluation prompts may not always be necessary. We further show that multilingual training improves performance in Spanish and Basque compared to strictly monolingual training, indicating beneficial cross-lingual transfer when learning the evaluation task. Interestingly, smaller 8B models perform competitively with larger 70B models, in some cases matching or surpassing proprietary alternatives, highlighting the potential of efficient open models for evaluation. However, in out-of-domain settings, fine-tuning on task-related out-of-domain data can degrade performance, and zero-shot evaluation often proves more robust.

Looking ahead, we plan to expand both the scale and diversity of our evaluation by collecting additional human-annotated data and developing new multilingual benchmarks, with a focus on low-resource languages. We also aim to explore downstream tasks to evaluate model generalization in varied real-world contexts. Together, these efforts will support the creation of more robust, transparent, and linguistically inclusive evaluation systems.

\section*{Limitations}
Our study is constrained primarily by the limited availability of suitable benchmarks, especially those that include Basque alongside major languages. The scarcity of multilingual evaluation resources restricts the diversity of tasks we can consider and may limit the generalizability of our findings across domains.

In addition, the training data used for our models is derived from a translated version of the \texttt{FeedBack Collection}, originally generated with GPT models. While this resource enables consistent and scalable experimentation across languages, it may not fully capture the variability, nuance, and pragmatic richness of naturally occurring human interactions. The absence of comparable human-authored datasets further highlights the current gap in high-quality, open multilingual resources for judgment and evaluation tasks. Similarly, as \texttt{RECON} and \texttt{FLASK} are machine-translated, the lack of natively generated benchmarks in Spanish and Basque limits our ability to capture language-specific pragmatics and may affect the extent to which our findings generalize to real-world use in these languages.

Finally, our study is limited to three languages, which, while selected to capture variation in both resource availability and linguistic typology, do not fully represent the diversity of multilingual settings. Expanding the analysis to additional languages would provide a more comprehensive picture.
\section*{Acknowledgments}
This work has been partially supported by the Basque Government (Research group funding IT-1805-22). We are also thankful to the following MCIN/AEI/10.13039/501100011033 projects: (i) DeepMinor (CNS2023-144375) funded by MTDFP/ and by European Union Next GenerationEU/ PRTR; (ii) DeepThought ((PID2024-159202OB-C21) and by ERDF, EU. Irune Zubiaga is supported by the UPV/EHU PIF24/08 predoc grant. We acknowledge the EuroHPC Joint Undertaking for awarding this project access to the EuroHPC supercomputer LEONARDO, hosted by CINECA (Italy) and the LEONARDO consortium through an EuroHPC Extreme Access call. We also acknowledge SCAYLE for granting this project access to the Calendula supercomputer through CLARIAH-ES.

\bibliography{custom}
\clearpage
\appendix
\section{Dataset Structure}\label{feedbackdataset}

All three datasets—\texttt{FeedBack Collection}, \texttt{RECON}, and \texttt{FLASK}—contain the following fields:

\begin{itemize}\setlength{\itemsep}{5pt}\setlength{\parskip}{0pt}\setlength{\parsep}{0pt}
\item \textbf{instruction}: The input provided to the evaluator model. It includes the instruction and response to be evaluated, the reference answer, and the scoring rubric.
\item \textbf{output}: The expected output from the evaluator model. It contains the feedback and the score decision, separated by the marker \texttt{[RESULT]}.
\item \textbf{orig\_instruction}: The original instruction to be evaluated. This differs from the full instruction, which includes additional context.
\item \textbf{orig\_response}: The model response being evaluated.
\item \textbf{orig\_reference\_answer}: The reference answer corresponding to the original instruction.
\item \textbf{orig\_criteria}: The scoring criteria used to evaluate the original response.
\item \textbf{orig\_score$x$\_description}: The description of conditions under which a score of \textit{x} (ranging from 1 to 5) should be assigned.
\end{itemize}

\section{Translation Prompt}\label{sec:prompt}
\begin{tcolorbox}[
                  breakable,
                  colback=gray!10,   
                  colframe=gray!80,  
                  boxrule=0.5pt,
                  left=2mm,
                  right=2mm,
                  top=1mm,
                  bottom=1mm]

You are a helpful AI assistant that specializes in English to \{target\_lang\} translations.
Your task is to translate instruction datasets from English to \{target\_lang\}.
\vspace{5pt}\\
Here are some important guidelines:
\vspace{5pt}\\
1. Maintain the original meaning and intent of the instructions
\vspace{1pt}\\
2. Use standard \{target\_lang\} language
\vspace{1pt}\\
3. Keep the technical terms that don't have widely accepted \{target\_lang\} translations
\vspace{1pt}\\
4. Preserve any code snippets, variables, or special characters exactly as they appear
\vspace{1pt}\\
5. Translate only the text content, not the JSON structure
\vspace{5pt}\\
The input will be a JSON object with English text. Please provide accurate \{target\_lang\} translations for all text fields.
\end{tcolorbox}



\section{Justification of the use of MT benchmarks}~\label{posvsmt}

Wilcoxon signed-rank tests~\cite{wilcoxon1945individual}, a nonparametric method for comparing two related samples, was used to evaluate whether the median differences between model predictions on the machine-translated (MT) and post-edited (PE) benchmarks were statistically significant. This test is appropriate when the data is paired but may not meet the normality assumption required by parametric tests. 
\begin{table}[htb]
\small
\centering
\resizebox{\columnwidth}{!}{%
\begin{tabular}{llrrr}
\hline
\textbf{Model} & \textbf{Lang} & \textbf{\textit{W}} & \textbf{\textit{p}} & \textbf{\textit{r}} \\
\hline
 DeepSeek Llama & \textit{multi} & 2214.5 & 0.677 & 0.019 \\
 & \textit{eu} & 10501.5 & 0.512 & 0.029 \\
 & \textit{io\_multi} & 671.0 & 0.215 & 0.055 \\
 & \textit{io\_eu} & 459.0 & 0.372 & 0.040 \\
\hline
 Latxa Instruct & multi& 3937.0 & 0.999 & 0.000 \\
 & eu & 3282.5 & 0.539 & 0.028 \\
 & io\_multi& 768.0 & 0.641 & 0.021 \\
 & io\_eu & 384.5 & 0.057 & 0.085 \\
 \hline
 Llama Instruct & multi & 3410.5 & 0.308 & 0.046 \\
 & eu & 5903.5 & 0.600 & 0.023 \\
 & io\_multi & 268.5 & 0.020$^*$ & 0.104 \\
 & io\_eu & 340.0 & 0.861 & 0.008 \\
\hline
\end{tabular}%
}
\caption{Results of the Wilcoxon signed-rank tests comparing model performance on the machine-translated (MT) and post-edited (PE) \texttt{RECON} benchmark across different language configurations. For each model and language setting, the table reports the Wilcoxon test statistic (\textit{W}), the associated p-value (\textit{p}), and the effect size (\textit{r}).}\label{tab:mt_comparison}
\end{table}

As shown in Table~\ref{tab:mt_comparison}, the results indicated no statistically significant differences between the MT and PE scores across models. The only exception was observed for the LLaMA Instruct model fine-tuned on io\_multi data, which showed a marginal uncorrected difference ($p = 0.020^*$) with a small effect size ($r = 0.104$), suggesting limited practical relevance. Effect sizes were interpreted following \citet{cohen1988statistical} conventions, where $r \approx 0.1$ denotes a small effect, $r\approx 0.3$ a medium effect, and $r \approx 0.5$ a large effect.

\begin{figure}[htb]
    \centering
    \includegraphics[width=\linewidth]{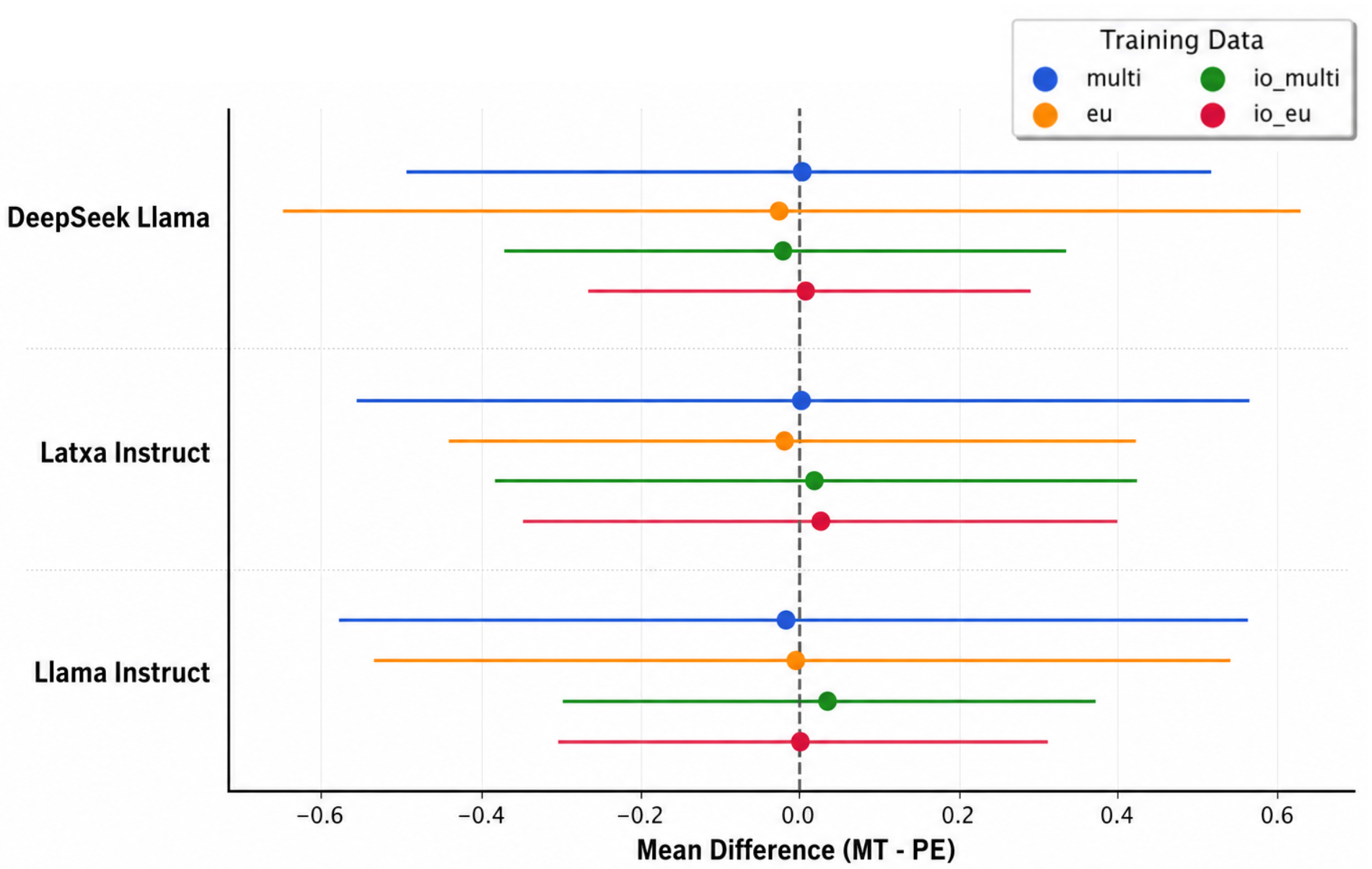}
    \caption{Mean differences between model scores on the machine-translated (MT) and post-edited (PE) \texttt{RECON} benchmark across language configurations. Each point represents the mean difference (MT – PE) for a given model and training data condition, while the horizontal lines indicate the standard deviation. Colors correspond to different training data settings (\textit{multi}, \textit{eu}, \textit{io\_multi}, and \textit{io\_eu}). The vertical dashed line at zero marks parity between MT and PE performance.}

    \label{fig:placeholder}
\end{figure}

To determine whether the pattern of MT–PE differences varied systematically across models, a Friedman test~\citep{friedman1937use} was conducted. This test evaluates whether multiple related samples differ consistently by comparing their ranked outcomes across conditions, and is suitable when the same models are tested under several related settings without assuming normality. The Friedman test yielded a non-significant result ($p = 0.498 > 0.05$), indicating that the null hypothesis could not be rejected. This suggests that the models did not differ systematically in their sensitivity to translation quality, and that the MT–PE performance differences were consistent across all seven models. The mean differences between model scores are reported in Figure~\ref{fig:placeholder}.

In summary, although using the machine-translated version may slightly reduce the precision with which model performance can be measured, the results indicate that the benchmark’s translation quality (machine-translated vs. post-edited) did not substantially affect the relative performance of the evaluated models, suggesting that the dataset is suitable for our objective of identifying the most effective training strategies for multilingual LLMs-as-a-judge.

\section{MT quality test}~\label{mt_quality}

To further assess the quality of the machine-translated data, in Table~\ref{tab:mt_quality} we report standard MT evaluation metrics (BLEU~\citep{papineni-etal-2002-bleu}, chrF~\citep{popovic-2015-chrf}, and COMET~\citep{rei-etal-2022-comet}) for \texttt{RECON} and \texttt{FLASK}. COMET is computed using the \texttt{wmt22-comet-da} model. For \texttt{RECON}, we compare machine-translated and post-edited instances, while for \texttt{FLASK}, we perform back-translation of the machine-translated data into English and compare it against the original source. The obtained scores are consistently high across metrics and language pairs, indicating strong translation quality.

\begin{table}[htp]
\centering
\small
\setlength{\tabcolsep}{8pt}
\resizebox{\columnwidth}{!}{%
\begin{tabular}{
l
S[table-format=2.4]
S[table-format=2.4]
S[table-format=2.4]
S[table-format=2.4]
}
\toprule

&
\multicolumn{2}{c}{\textbf{RECON}} &
\multicolumn{2}{c}{\textbf{FLASK}} \\

\cmidrule(lr){2-3}
\cmidrule(lr){4-5}

\textbf{Metric} &
{\textbf{ES}} &
{\textbf{EU}} &
{\textbf{ES}} &
{\textbf{EU}} \\

\midrule

BLEU  & 0.8729 & 0.6270 & 0.7566 & 0.6551 \\
chrF  & 92.8026 & 85.4385 & 85.1522 & 78.1205 \\
COMET & 0.9109 & 0.9144 & 0.9038 & 0.8911 \\

\bottomrule
\end{tabular}
}

\caption{Machine translation quality evaluation using BLEU, chrF, and COMET scores. RECON compares machine translated instances with post-edited references, while FLASK evaluates back-translation against the original English source.}
\label{tab:mt_quality}
\end{table}

\section{Label Extraction Pattern}~\label{reg_pat}
We use the following regular expression to extract the judgment labels from the text:
\begin{verbatim}
.{90,}\[(RESULT|EMAITZA|RESULTADO)\]\s*
[1-5]
\end{verbatim}

This pattern matches sequences of at least 90 characters followed by a language-specific tag (RESULT, EMAITZA, or RESULTADO) and a score from 1 to 5 at the end of the line. It is applied both during guided decoding in vLLM to constrain the model outputs and afterwards to verify the extracted labels.


\section{Multilingual training and cross-lingual transfer}~\label{ap:multiandcross}

\begin{table*}[htbp]
\centering
\small
\setlength{\tabcolsep}{6pt}

\begin{tabular}{
l l
S[table-format=1.3]
S[table-format=1.3]
S[table-format=1.3]
S[table-format=1.3]
S[table-format=1.3]
S[table-format=1.3]
}
\toprule

& &
\multicolumn{2}{c}{\textbf{English (en)}} &
\multicolumn{2}{c}{\textbf{Spanish (es)}} &
\multicolumn{2}{c}{\textbf{Basque (eu)}} \\

\cmidrule(lr){3-4}
\cmidrule(lr){5-6}
\cmidrule(lr){7-8}

\textbf{Model} &
\textbf{Train Lang} &
{$r \uparrow$} &
{MSE $\downarrow$} &
{$r \uparrow$} &
{MSE $\downarrow$} &
{$r \uparrow$} &
{MSE $\downarrow$} \\

\midrule

\multirow{4}{*}{DeepSeek-R1-Distill-Llama-8B}
& en        & 0.838 & 0.806 & 0.766 & 1.200 & 0.586 & 1.818 \\
& io\_es    & 0.784 & 1.002 & 0.787 & 1.058 & 0.611 & 1.842 \\
& io\_eu    & 0.803 & 0.926 & 0.766 & 1.172 & 0.728 & 1.246 \\
& io\_multi & 0.802 & 0.932 & 0.783 & 1.014 & 0.760 & 1.060 \\

\midrule

\multirow{4}{*}{Latxa3.1\_Instruct\_8B}
& en        & 0.859 & 0.712 & 0.775 & 1.064 & 0.765 & 1.138 \\
& io\_es    & 0.795 & 0.928 & 0.787 & 0.994 & 0.754 & 1.192 \\
& io\_eu    & 0.738 & 1.130 & 0.759 & 1.048 & 0.761 & 1.010 \\
& io\_multi & \bfseries 0.836 & \bfseries 0.724 &
               \bfseries 0.816 & \bfseries 0.824 &
               \bfseries 0.805 & \bfseries 0.888 \\

\midrule

\multirow{4}{*}{Llama-3.1-8B-Instruct}
& en        & 0.825 & 0.802 & 0.447 & 4.130 & 0.556 & 2.122 \\
& io\_es    & 0.722 & 1.246 & 0.740 & 1.160 & 0.649 & 1.664 \\
& io\_eu    & 0.723 & 1.216 & 0.747 & 1.132 & 0.683 & 1.346 \\
& io\_multi & 0.741 & 1.102 & 0.757 & 1.044 & 0.732 & 1.140 \\

\bottomrule
\end{tabular}

\caption{Cross-lingual performance across English (en), Spanish (es), and Basque (eu). Higher Pearson correlation ($r$) and lower mean squared error (MSE) indicate better performance.}
\label{tab:crosslingual_results}

\end{table*}
\end{document}